\documentclass[10pt,twocolumn,letterpaper]{article}

\usepackage{iccv}
\usepackage{times}
\usepackage{epsfig}
\usepackage{graphicx}
\usepackage{amsmath}
\usepackage{amssymb}
\usepackage{pbox}

\usepackage{color}
\usepackage{amsmath}
\usepackage{amssymb}
\usepackage{multirow, varwidth}
\usepackage{verbatim}
\makeatletter
\newif\if@restonecol
\makeatother

\usepackage[ruled,vlined]{algorithm2e}
\usepackage{xr}
\usepackage[amsmath,thmmarks]{ntheorem}

\theorembodyfont{\normalfont}
\theoremseparator{.}
\theoremstyle{nonumberplain}

\DeclareRobustCommand\onedot{\futurelet\@let@token\@onedot}
\def\onedot{.\@\xspace}

\def\eg{\emph{e.g}\onedot} 
\def\ie{\emph{i.e}\onedot}

\def\etal{\emph{et al}\onedot}

\usepackage{color}
\usepackage{etoolbox}

\newcommand*{\affaddr}[1]{#1} % No op here. Customize it for different styles.
\newcommand*{\affmark}[1][*]{\textsuperscript{#1}}

\usepackage[pagebackref=true,breaklinks=true,letterpaper=true,colorlinks,bookmarks=false]{hyperref}

\DeclareGraphicsExtensions{.pdf,.png,.jpg}

\usepackage[breaklinks=true,bookmarks=false]{hyperref}

\iccvfinalcopy % *** Uncomment this line for the final submission

\def\iccvPaperID{164} % *** Enter the ICCV Paper ID here

\makeatletter
  \let\ps@plain\ps@empty
\makeatother
\patchcmd{\chapter}{plain}{empty}{}{}

\begin{document}

%%%%%%%%% TITLE
\title{A Read-Write Memory Network for Movie Story Understanding}

\author{%
Seil Na\affmark[1], Sangho Lee\affmark[1], Jisung Kim\affmark[2], Gunhee Kim\affmark[1] \\
\affaddr{\affmark[1]Seoul National University, \affmark[2]SK Telecom}\\
{\tt\small \{seil.na, sangho.lee\}@vision.snu.ac.kr, joyful.kim@sk.com, gunhee@snu.ac.kr} \\
\tt\small \url{https://github.com/seilna/RWMN}
}

\maketitle

%%%%%%%%% ABSTRACT
\begin{abstract}

%Recent studies for memory augmented neural network models have focused on how to systematically manage and exploit memories to improve the performance of models for various AI tasks such as visual question and answering (QA). 
We propose a novel memory network model named Read-Write Memory Network (RWMN)  to perform question and answering  tasks for large-scale, multimodal movie story understanding.
The key focus of our RWMN model is to design the \textit{read network} and the \textit{write network} that consist of multiple convolutional layers, 
which enable memory read and write operations to have high capacity and flexibility. 
While existing memory-augmented network models treat each memory slot as an independent block,
our use of multi-layered CNNs allows the model to read and write sequential memory cells as chunks,
which is more reasonable to represent a sequential story because adjacent memory blocks often have strong correlations. 
For evaluation, we apply our model to all the six tasks of the MovieQA benchmark~\cite{tapaswi2016movieqa}, and achieve the best accuracies on several tasks, especially on the visual QA task.
Our model shows a potential to better understand not only the content in the story, but also more abstract information, such as relationships between characters and the reasons for their actions. 
\end{abstract}

%%%%%%%%% BODY TEXT
\section{Introduction}

%%%%%%% Intro
%------------------------------------------------------------------------------
% Figure 1: Key idea
\begin{figure}[t]
\centering
\includegraphics[trim=0.2cm 0.2cm 0cm 0.1cm,clip,width=0.47\textwidth]{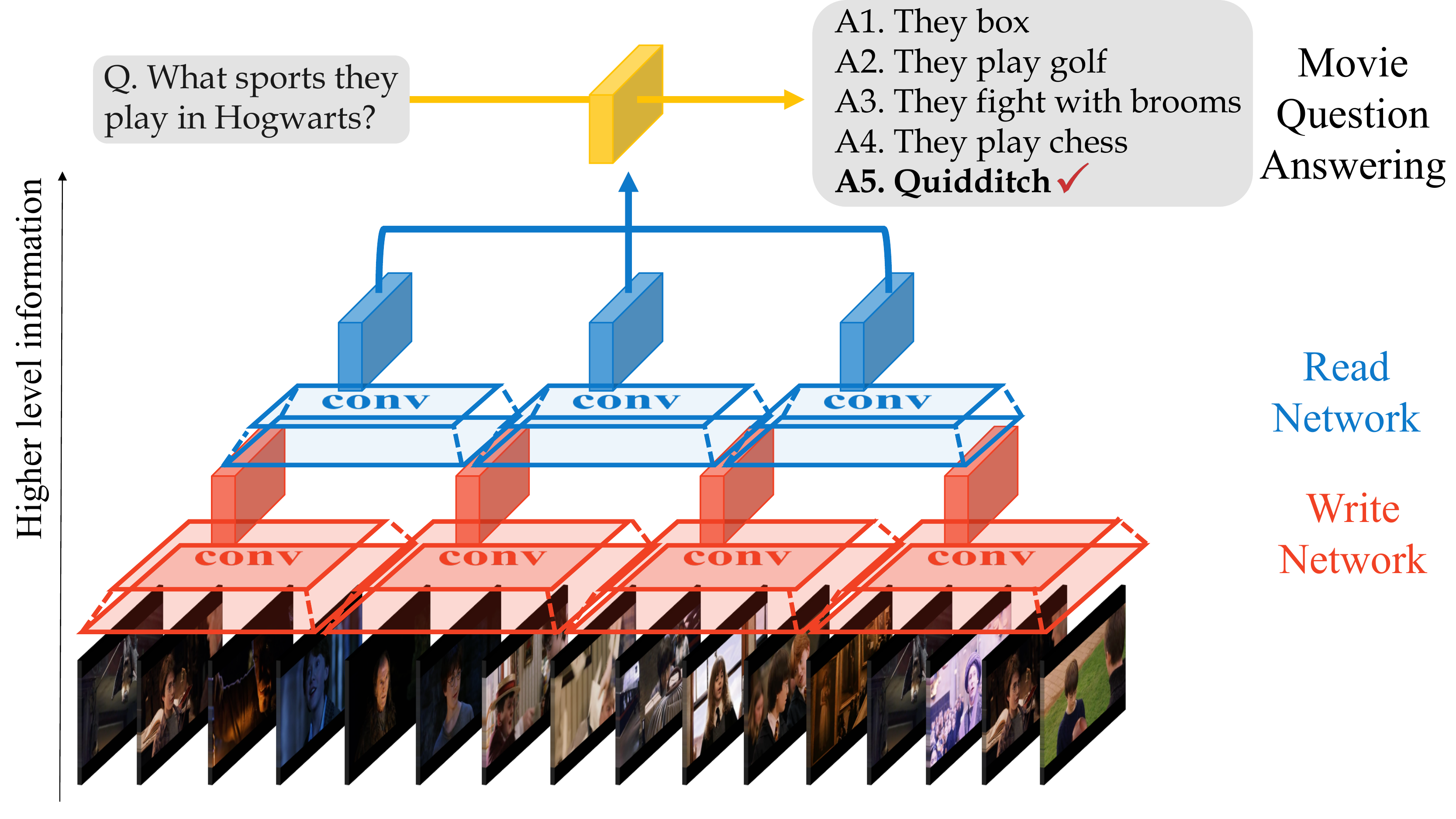}
%\vspace{-10pt}
\caption{The intuition of the RWMN (\textit{Read-Write Memory Network}) model for movie question and answering tasks. 
Using read/write networks of multi-layered CNNs, it abstracts a given series of frames stepwise to capture higher-level sequential information and stores it into memory slots. It eventually helps answer complex questions of movie QAs.
%Memory blocks with high level of abstraction information is great help to 
}
\label{fig:overview}
\end{figure}
%------------------------------------------------------------------------------
For many problems of video understanding, including video classification~\cite{abu2016youtube,karpathy2014l}, video captioning~\cite{xu2016msr,yu2017lsmdc} and MovieQA~\cite{tapaswi2016movieqa}, it is key to success for models to correctly process, represent, and store long sequential information. 
% are the tasks where it is key to success to correctly process the input of sequential data. 
In the era of deep learning, one prevailing approach to model sequential input is to use recurrent neural networks  (RNNs)~\cite{mikolov2010recurrent} which store the given information into a hidden memory and update it over time. %, as an embedding of the whole sequential input. 
However, RNNs accumulate information in a single fixed-length memory regardless of the length of an input sequence, thus tend to fail to utilize far-distant information due to a vanishing gradient problem, which is still not fully solved even with advanced models such as LSTM~\cite{hochreiter1997long} and GRU~\cite{cho2014learning}.

As another recent alternative to resolve this issue, many studies attempt to  leverage an external memory structure for neural networks, often referred to as \textit{neural memory networks}~\cite{graves2014neural,graves2016hybrid,gulcehre2016dynamic,kumar2015ask,sukhbaatar2015end,weston2014memory}.
One key benefit of external memory is to enable a neural model to cache sequential inputs in memory slots, and explicitly utilize even far early information. 
Such ability is particularly powerful to solve \textit{question and answering} (QA) problems, which often require models to memorize a large amount of information, and correctly access the most relevant information to a given question. 
For this reason, memory networks have been popularly applied as state-of-the-art approaches to many QA tasks, such as bAbI task~\cite{weston2015towards}, SQuAD~\cite{rajpurkar2016squad}, and LSMDC~\cite{rohrbach2016movie}.

MovieQA~\cite{tapaswi2016movieqa} is another challenging visual QA dataset, in which models need to understand movies over two hours long, and solve QA problems related to movie content and plots. 
The MovieQA benchmark consists of six tasks according to which sources of information is usable to solve the QA problems, including videos, subtitles, DVS, scripts, plot synopses, and open-end information. 
Understanding a movie is a highly challenging task; it is necessary not only to understand the content of individual video frames such as a characters' actions, places of events, but also to infer more abstract and high-level knowledge such as reasons of a characters' behaviors, and relationships between them. 
% Furthermore, one of the ultimate goals of machine learning, "artificial intelligence in the real world," requires a similarly complex and abstract understanding. 
For instance, in the \textit{Harry Potter} movie, to answer a question (\textit{Q. What does Harry trick Lucius into doing? A. Freeing Dobby}), models need to realize that \textit{Dobby was a Lucius's house elf, wanted to escape from him, had a positive relationship with Harry, and Harry helped him}. 
Some of such information is visually or textually observable in the movie, but much information like relationships between characters and correlations between events should be deduced. 
% In summary, understanding a long multimodal story, such as a movie, may require a higher level of reasoning than existing QA tasks. 

Our objective is to propose a novel memory network model to perform QA tasks for large-scale, multimodal movie story understanding. 
That is, the input to the model can be very long (\eg videos more than two hours long), or be multimodal (\eg text-only or video-text pairs). 
The key focus of our novel memory network named \textit{Read-Write Memory Networks} (RWMN)  is on defining the memory read/write operations to have high capacity and flexibility, 
for which we propose the \textit{read} and \textit{write networks} that consist of multiple convolutional layers. 
Existing neural memory network models treat each memory slot as an independent block. 
However, adjacent memory blocks often have strong correlations, which are the case to represent a sequential story. 
That is, when human understands a story, the entire story is often recognized as a sequence of closely-interconnected abstract events. %, rather than individual sentences or scenes.
Hence, preferably memory networks need to read and write sequential memory cells as chunks, which are implemented by multiple convolutional layers of the read and write network.

To conclude introduction, we summarize the contributions of this work as follows.

\begin{enumerate}
\item We propose a novel memory network named RWMN  that enables the model to flexibly read and write more complex and abstract information into memory slots through read/write networks. 
To the best of our knowledge, it is the first attempt to leverage multi-layer CNNs for read/write operations of a memory network. 

\item The RWMN shows the best accuracies on several tasks of MovieQA benchmark~\cite{tapaswi2016movieqa}; as of the ICCV2017 submission deadline (March 27, 2017 23:59 GMT), our RWMN achieves the best performance for \textit{four} out of five tasks in the validation set, and \textit{four} out of six tasks in the test set.
Our quantitative and qualitative evaluation also assures that the read/write networks effectively utilize higher-level information in the external memory, especially on the visual QA task. %(as of ICCV submission) , 
% We observe this way of constructing memory is more advantageous to access the stories in the memory as higher level events or episodes.
% From this viewpoint of the model, it can be said that, rather than constructing a memory slot on a sentence-by-sentence basis, 
% We evaluate the RWMN for several tasks of MovieQA public datasets, and show that it performs better than existing approaches in five categories and that models can catch abstract information well from qualitative results. 
%To validate the applicability of the proposed approach, we participate in all the four tasks of LSMDC 2016.
%Our models achieve the best accuracies in three of them, including \textit{fill-in-the-blank}, \textit{multiple-choice test}, and \textit{movie retrieval}.
%We also attain comparable performance for the other task \textit{movie description}.
\end{enumerate}

%------------------------------------------------------------------------
%%%%%%%%%%% Related work.
\section{Related Work}
\label{sec:related_work}

\textbf{Neural Memory Networks}.
Recently, much research has been done to model sequential data using explicit memory architecture. 
The memory access of existing memory network models can be classified into \textit{content-based} addressing and \textit{location-based} addressing~\cite{graves2014neural}. 
% such as storing the content of the story in memory in a more systematic manner and appropriately abstracting stored information to answer a given question. 
The content-based addressing (\eg\cite{graves2016hybrid,weston2014memory, miller2016key}) lets the controller to generate a key vector and measure its similarity with each memory cell to find out which cells are to be \textit{attended} as the relevant cells to the key vector.
Location-based addressing (\eg\cite{graves2014neural}), on the other hand, enables simple arithmetic operations that find out the addresses to store or retrieve information, regardless of the content of the key vector.

Neural Turing Machine (NTM)~\cite{graves2014neural} and its extensions of DNC~\cite{graves2016hybrid}, D-NTM~\cite{gulcehre2016dynamic}, focus on learning the entire process of memory interaction (read/write operations), and thus the degree of freedom (or capability) of the model is high in solving a given problem.  
They have been successfully applied to complex tasks such as sorting, sequence copying, and graph traversal.
% our model are designed to organize memory specifically to understand a given story and to extract information about a question from it.
The memory networks of \cite{kumar2015ask,sukhbaatar2015end,weston2014memory} address the QA problems using continuous memory representation similar to the NTM. 
However, while the NTM leverages both content-based and location-based addressing, they use only the former (content-based) memory interaction. % with the similar structure. 
They apply the concept of multi-hops to recurrently read the memory, which results in performance improvement in solving QA problems that require causal reasoning. 
The work of \cite{miller2016key,zhang2016dynamic} proposes a key-value memory network that stores information in the form of (key, value) pairs into the external knowledge base. These methods are good at solving QA problems that  focus on the content or facts in a context such as WikiMovies~\cite{miller2016key} and bAbI dataset~\cite{weston2015towards}.

The work of \cite{chandar2016hierarchical,rae2016scaling} deals with how to make the read/write operations scalable with extremely large amount of memory. Chandar \etal\cite{chandar2016hierarchical} propose to organize memory hierarchically, and Rae \etal\cite{rae2016scaling} make read and write operations sparse, thereby increasing scalability and reducing the cost of operations.  Cesc \etal\cite{attend2u:2017:CVPR} adopt convolutional read from memory to jointly represent nearby ordered memory slots.

Compared to all the previous models, our RWMN model is explicitly equipped with learnable read/write networks of CNNs,
which are specialized in storing and utilizing more abstract information, such as relationships between characters, reasons for characters' specific behaviors, as well as understanding of facts in a given story.
% The applied by them is based on reasoning about the fact in a given context. In order to understand a long story like Movie, it is necessary to focus on the "flow" of the story and the interaction between the characters beyond facts. Thus, based on these studies, 
% These QA problems are designed to understand and to measure the understanding of the model for the context.

\smallskip
\textbf{Models for MovieQA}.
Among the models applied to the MovieQA benchmark~\cite{tapaswi2016movieqa}, the end-to-end memory network~\cite{sukhbaatar2015end} is the state-of-the-art approach. It splits each movie into shot subshots, and constructs memory slots with video and subtitle features. 
It then uses content-based addressing to attend on the information relevant to a given question. 
Recently, Wang and Jiang~\cite{wang2017compare} present the compare-aggregate framework for word-level matching to measure the similarity of sentences. 
However, it is applied to only a single task (plot synopses) of MovieQA. 

There have been also several studies to solve Video QA tasks in other datasets, such as LSMDC~\cite{rohrbach2016movie}, MSR-VTT~\cite{xu2016msr}, and TGIF-QA~\cite{jang2017tgifqa}, which mainly focus on understanding short video clips, and answering about factual elements in the clips. Yu \etal~\cite{yu2017lsmdc} achieve compelling performance in video captioning, video QA, and video retrieval by constructing an end-to-end trainable concept-word-detector along with vision-to-language models.

%------------------------------------------------------------------------
% Figure 2. RWMN model.

\section{ Read-Write Memory Network (RWMN)}
\label{sec:rwmn}

\begin{figure*}
\begin{center}
    \includegraphics[width=0.99\textwidth]{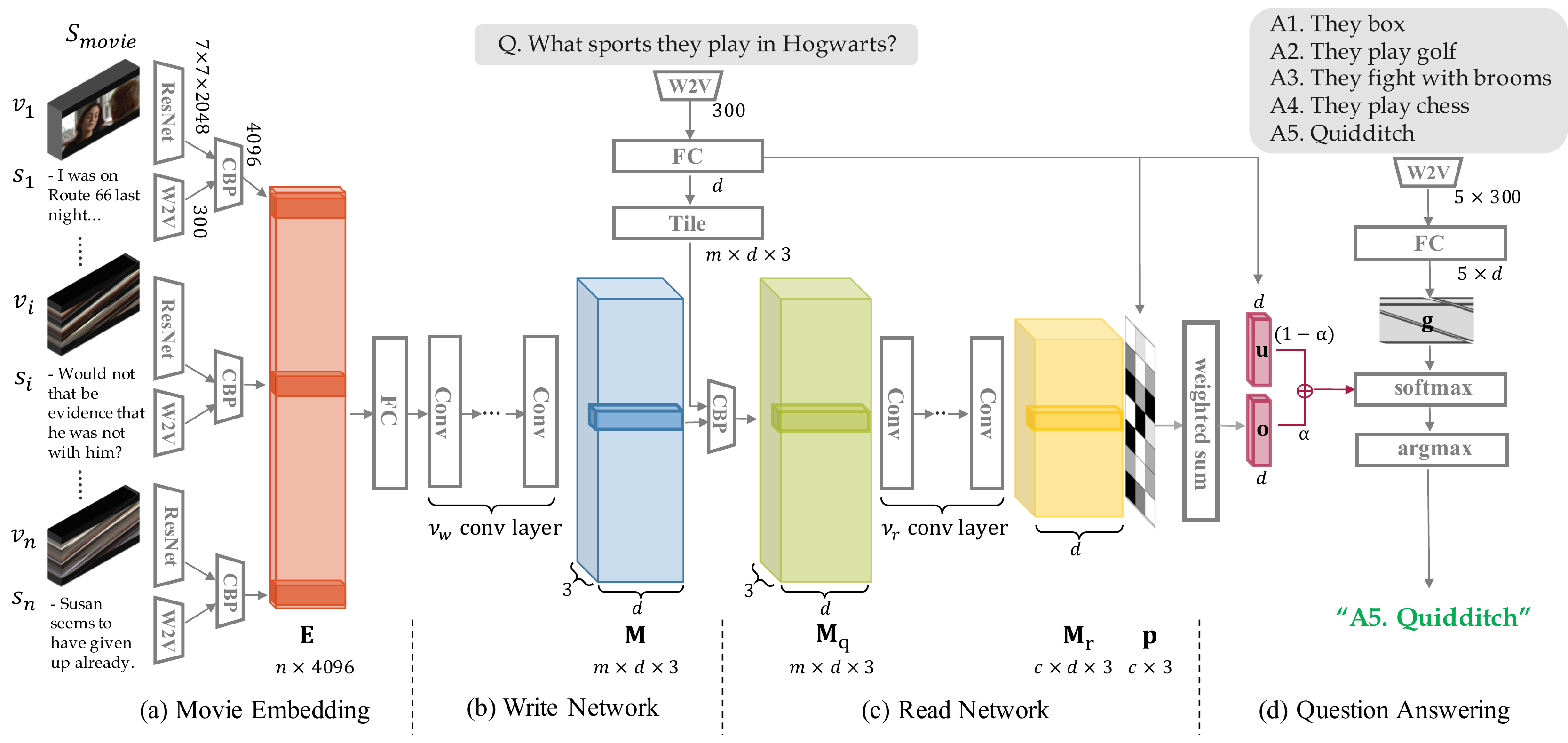}
\end{center}
   \caption{Illustration of the proposed \textit{Read-Write Network}. 
   (a) The multimodal movie embedding $\mathbf{E}$ is obtained using the ResNet feature and the Word2Vec representation from movie subshots and subscripts (section \ref{sec:movie_embedding}). 
   (b) The write memory $\mathbf M$ abstracts higher-level sequential information through multiple convolution layers (section \ref{sec:write_net}). 
   (c) The query-dependent memory $\mathbf M_q$ is obtained via the Compact Bilinear Pooling (CBP) between the query and each slot of $\mathbf M$, and then the read memory $\mathbf M_r$ is constructed through convolution layers (section \ref{sec:read_net}).
   (d) Finally, the answer with the highest confidence score is chosen out of five candidates (section \ref{sec:answer_selection}).
}
\label{fig:RWMN}
\end{figure*}

Figure \ref{fig:RWMN} shows the overall structure of our RWMN.
The RWMN is trained to store the movie content with proper representation in the memory, 
extract relevant information from memory cells in response to a given query, and select correct answer from five choices.

Based on the QA format of MovieQA dataset \cite{tapaswi2016movieqa}, the input of the model is (i) a sequence of video segment and subtitle pairs $S_{movie}= \{(v_1, s_1), ..., (v_n,s_n)\}$ for the whole movie, which takes about 2 hours ($n \sim 1,558$ on average), (ii) a question $q$ for the movie, and (iii) five answer candidates $a = \{a_1, ... , a_5\}$.
In the video+subtitle task of MovieQA, for example, each $s_i$ is a dialog sentence of a character, and $ v_i = \{v_{i1}, ..., v_{im}\}$ is a video subshot (\ie a set of frames)  sampled at 6 fps that are temporally aligned with $s_i$. 
The output is a confidence score vector over the five answer candidates.

In the following, we explain the architecture according to information flow, from movie embedding to answer selection via write/read networks. 

\subsection{Movie Embedding}
\label{sec:movie_embedding}

We convert each subshot $v_i$  and text sentence $s_i$ into feature representation as follows. % and use it to create movie embedding $\mathbf{E}$.
For each frame $v_{ij} \in v_i$, we first obtain its feature $\mathbf v_{ij}$ by applying the ResNet-152~\cite{he2016deep} pretrained on ImageNet~\cite{deng2009imagenet}. 
We then mean-pool over all frames as $\mathbf v_{i} = \sum_{j} \mathbf v_{ij} \in \mathbb{R}^{7 \times 7 \times 2,048}$, as a representation of the subshot $v_{i}$.
For each sentence $s_i$, we first divide the sentence into words, apply the pretrained Word2Vec~\cite{mikolov2013distributed}, and then mean-pool with the position encoding (PE)~\cite{sukhbaatar2015end} as $\mathbf s_{i} = \sum_{j}\mbox{PE}(\mathbf s_{ij}) \in \mathbb{R}^{300}$.

Finally, to obtain a multimodal space embedding of $\mathbf v_i$ and $\mathbf s_i$, we use the Compact Bilinear Pooling (CBP)~\cite{fukui2016multimodal} as 
\begin{equation}
\label{eq:cbp}
\mathbf{E}[i] = \mbox{CBP}(\mathbf v_i,  \mathbf s_i) \in \mathbb{R}^{4,096}.
\end{equation}
We perform this procedure for all $n$ pairs of subshots and text, resulting in a 2D movie embedding matrix $\mathbf{E} \in \mathbb{R}^{n \times 4,096}$,
which is the input of our \textit{write network}.

\subsection{The Write Network}
\label{sec:write_net}

The write network takes a movie embedding matrix $\mathbf{E}$ as an input 
and generates a memory tensor $\mathbf{M}$ as output. 
The write network is motivated by that when human understands a movie, she does not remember it as a simple sequence of speech and visual content, 
but rather ties together several adjacent utterances and scenes in a form of events or episodes.
That is, each memory cell needs to associate neighboring movie embeddings, instead of storing each of $n$ movie embedding separately. 
To implement this idea of jointly storing adjacent embeddings into every slot, we exploit a convolutional neural network (CNN) as the write network. 
We experimentally confirm the following CNN design after thorough tests, by varying the dimensions, depths, strides of convolution layers. 

To the movie embedding $\mathbf{E} \in \mathbb R^{n\times4,096}$, 
we first apply a fully connected (FC) layer with parameter $\mathbf{W}_{c} \in \mathbb R^{4,096 \times d}, \mathbf{b}_{c} \in \mathbb{R}^{d}$ to project each $\mathbf{E}[i]$ into a $d$-dimensional vector.
The FC layer reduces the dimension of $\mathbf{E}$ in order to equalize the dimensions of query embedding and answer embedding, 
which is also beneficial to reduce the number of required convolution operations later.
We then use a convolution layer consisting of a filter $\mathbf w^w_{conv} \in \mathbb R^{f^w_v\times f^w_h \times 1 \times f^w_c}$, whose vertical and horizontal filter size is $f^w_v=40, f^w_h=d$, the number of filter channel is $f^w_c=3$ and strides are $s^w_v=30$ and $s^w_h=1$, respectively: 
\begin{equation} 
\label{eq:writeM}
\mathbf{M} = \mbox{ReLU}( \mbox{conv} ( (\bold{E} \mathbf{W}_{c} + \bold{b}_{c}) , \mathbf{w}^w_{conv}, \mathbf{b}_{w}))
\end{equation}
\noindent where conv (input, filter, bias) indicates the convolution layer, $\mathbf{b}_{w} \in \mathbb{R}^{f^w_c}$ is a bias, and ReLU indicates the element-wise ReLU activation~\cite{nair2010rectified}. 
Finally, the generated memory is $\mathbf{M} \in \mathbb{R}^{m \times {d} \times 3} $, where $m=\left \lfloor{({(n-1)/s^w_v} + 1)}\right \rfloor$. 

Note that the write network can employ multiple convolutional layers. If the number of layers is $\nu_w$, then we obtain $\mathbf{M}$ by recursively applying 
\begin{equation} 
\label{eq:writeMmult}
\mathbf{M}^{(l+1)} = \mbox{ReLU}( \mbox{conv}  (\bold{M}^{(l)}, \mathbf{w}^{w(l)}_{conv}, \mathbf{b}_{w}^{(l)}))
\end{equation}
\noindent from $l=1\ldots,\nu_w-1$.
In section \ref{sec:experiments}, we will report the result of ablation study to find out the best-performing $\nu_w$.

\subsection{The Read Network}
\label{sec:read_net}

The \textit{read network} takes a question $q$ and then generate answer from a compatibility between $q$ and $\mathbf M$.

\smallskip
\textbf{Question embedding}. 
We embed the question sentence $q$ as follows. 
We first obtain the Word2Vec vector~\cite{mikolov2013distributed} $\mathbf q$ as done in section \ref{sec:movie_embedding}, and then project it as follows.
\begin{equation}
\label{eq:questioin_embedding}
\mathbf u = \mathbf W_q \mathbf q + \mathbf b_q
\end{equation}
\noindent where parameters are $\mathbf W_q  \in \mathbb{R}^{d \times 300}$ and $\mathbf b_q \in \mathbb{R}^{d}$.

% First, $q$, which means one question sentence, is transformed into \textcolor{blue}{$d$}-dimension embedding $\mathbf{u}$ %like $s_i$ and 
Next the read network takes the memory $\mathbf{M}$ and the query embedding $\mathbf{u}$ as input, and generates the confidence score vector $\mathbf{o} \in \mathbb{R}^d$
as follows. 
% Then, when generating output vector, our Read Network focuses on (1) query-dependent memory modification and (2) memory reconstruction.

\smallskip
\textbf{Query-dependent memory embedding}.
We first transform the memory $\mathbf{M}$ to be query-dependent. 
Its intuition is that, according to the query, different types of information must be retrieved from the memory slots. 
For example, for the \textit{Harry Potter} movie, suppose that one memory slot contains the information about a particular scene where \textit{Harry is chanting magic spells}.
This memory slot should be read differently according to two different questions $Q_1$: \textit{What color is Harry wearing?} and $Q_2$: \textit{Why is Harry chanting magic spells?}
In section \ref{sec:experiments}, we will empirically show the effectiveness of this question-dependent memory update.

To transform the memory $\mathbf{M}$ into a query-dependent memory $\mathbf{M}_q \in \mathbb{R}^{m \times d \times 3}$, 
we apply the CBP~\cite{fukui2016multimodal} between each memory cell of $\mathbf{M}$ and the query embedding $\mathbf{u}$ as
\begin{equation}
\label{eq:queryM}
\mathbf M_q [i,:,j] = \mbox{CBP}(\mathbf{M}[i,:,j], \mathbf{u})
\end{equation}
\noindent for all $i=1,\cdots,m$, and $ j=1,2,3$.

%As a result, output containing higher-level information is obtained than content-based addressing, which obtains an output vector by assigning different attention to each fixed memory cell.
%To create an output vector $\mathbf{o}$, we need to get information from the memory we have configured.

\smallskip
\textbf{Convolutional memory read}.
% To generate an output vector from the modified memory, we go through a \textit{memory reconstruction} process.
As done in the write network, we also leverage a CNN to implement the read network. 
Our intuition is that, for correctly answering the question of movie understanding, it is important to connect and relate a series of scenes as a whole. 
Therefore, we use the CNN architecture to access chunks of sequential memory slots. %, rather than individual memory slots.
We obtain the reconstructed memory $\mathbf M_r$ by applying convolutional layers with a filter $\mathbf w^r_{conv} \in \mathbb{R}^{f^r_v \times f^r_h \times 3 \times f^r_c}$ 
whose vertical and horizontal filter size is $f^r_v=3, f^r_h=d$, the number of filter channel is $f^r_c=3$ and strides are $s^r_v=1$, $s^r_h=1$, respectively.
Finally, the reconstructed memory is $\mathbf{M}_r \in \mathbb{R}^{c \times {d} \times 3} $ with  $c=\left \lfloor{  (m-1)/s^r_v} + 1 \right \rfloor$:
\begin{equation}
\mathbf M_r = \mbox{ReLU}( \mbox{conv} ( \mathbf M_q , \mathbf{w}^r_{conv}, \mathbf{b}_{r}))
\label{eq:readM}
\end{equation}
\noindent where $\mathbf b_r \in \mathbb{R}^3$ is a bias term.
As in the write network, the read network can also have a $\nu_r$ number of stacks of convolutional layers; the formulation is the same with Eq.(\ref{eq:writeMmult}) only except replacing $\mathbf M, \mathbf{w}^w_{conv}, \mathbf{b}_{w}$  with $\mathbf M_r, \mathbf{w}^r_{conv}, \mathbf{b}_{r}$, respectively.
We will also report the results of ablation study about different $\nu_r$ in section \ref{sec:experiments}.

\subsection{Answer Selection}
\label{sec:answer_selection}

Next we compute the attention matrix $\mathbf{p} \in \mathbb{R}^{c \times 3}$ through applying the softmax to the dot product between the query embedding $\mathbf{u}$ and each cell of memory $\mathbf M_r$:
\begin{equation}
\mathbf{p}[i,j] = \mbox{softmax}(\mathbf M_r[i,:,j] \cdot \mathbf u )
\label{eq:read_attention}
\end{equation}
\noindent where $\cdot$ indicates the dot product. 
Finally, the output vector $\mathbf{o} \in \mathbb{R}^d$ is obtained through a weighted sum between each memory cell of $\mathbf M_r$ and the attention vector $\mathbf{p}$:
\begin{equation}
\label{eq:output_vector}
\mathbf{o} [i] = \sum_{j=1}^{c} \sum_{k=1}^3 \mathbf M_r [j,i,k] \mathbf{p}[j,k].
\end{equation}
Next we obtain the embedding of five answer candidate sentences $\{a\}$ as done for the question in Eq.(\ref{eq:questioin_embedding}) with sharing the parameters $\mathbf W_q$ and $\mathbf b_q$. As a result, we compute the embedding of answer candidates $\mathbf{g} \in \mathbb{R}^{5 \times d}$. 

We compute the confidence vector $\mathbf{z} \in \mathbb{R}^{5}$ by finding the similarity between $\mathbf{g}$ and the weighted sum of $\mathbf{o}$ and $\mathbf{u}$.
\begin{equation}
\mathbf{z} = \mbox{softmax}( (\alpha \mathbf o + (1 - \alpha) \mathbf u)^T \mathbf{g}), 
\end{equation}
where $\alpha \in [0, 1]$ is a trainable parameter.
Finally, we predict the answer $y$ with the highest confidence score: $y = \mbox{argmax}_{i \in [1,5]} (\mathbf{z}_i)$.

%----------------------------------------------------------------------------------------
%%%%%%% Training
\subsection{Training}
\label{sec:training}

For training of our model, we minimize the softmax cross-entropy between the prediction $\mathbf{z}$ and the groundtruth one-hot vector $\mathbf{z}_{gt}$.
All training parameters are initialized with the Xavier method~\cite{glorot2010understanding}.
%We apply a stochastic gradient descent with a mini-batch size of 32.
Experimentally, we select the Adagrad~\cite{duchi2011adaptive} optimizer with a mini-batch size of 32, a learning rate of 0.001, and an initial accumulator value of 0.1.
We train our model up to 200 epochs, although we actively use the early stopping to avoid overfitting due to the small size of the MovieQA dataset.
We repeat training each model with 12 different random initializations, and select the one with the lowest cost. 
%We observe that the performance of the model shows variance according to initialization. Hence, we repeat training 12 times for the same model with different initializations and select the one with the lowest cost.

%-------------------------------------------------------------------------------------------
%%%%%%%%% Experiments
\section{Experiments}
\label{sec:experiments}

We evaluate the proposed RWMN model for all the tasks of MovieQA benchmark~\cite{tapaswi2016movieqa}. 
We defer more experimental results and implementation details to the supplementary file.
% We plan to make our source code public.

%%%%%%%%% 4.1 Task description
\subsection{MovieQA Tasks and Experimental Setting}
\label{sec:movieqa_dataset}

%%%% Table 1. MovieQA Dataset statistic.
\begin{table}[t]
\small
\begin{center}
\begin{tabular}{l|cc}
\hline
Story sources & \# movie & \# QA pairs \\
\hline
Videos and subtitles & 140 & 6,462 \\
Subtitles & 408 & 14,944 \\
DVS & 60 & 2,446\\
Scripts & 199 & 7,810\\ 
Plot synopses & 408 & 14,944 \\
\hline
\end{tabular}
\vspace{6pt}
\caption{The number of movies and QA pairs according to data sources in the MovieQA dataset~\cite{tapaswi2016movieqa}.}
\vspace{-13pt}
\label{tab:dataset}
\end{center}
\end{table}

As summarized in Table \ref{tab:dataset}, MovieQA dataset~\cite{tapaswi2016movieqa} contains 408 movies and 14,944 multiple choice QA pairs, each of which consists of five answer choices with only one correct answer.
The dataset provides with five types of story sources associated with the movies: videos, subtitles, DVS, scripts, and plot synopses, 
based on which the MovieQA challenge hosts 6 subtasks, according to which sources of information are differently used:
(i) video+subtitle, (ii) subtitles only, (iii) DVS only, (iv) scripts only, (v) plot synopses only, and (vi) open-ended. 
That is, there are one video-text QA task, and four text-only QA tasks, and one open-end QA task  with no restriction on additional story sources. 
We strictly follow the test protocols of the challenge, including training/validation/test split and evaluation metrics. 
More details of the dataset and rules are available in \cite{tapaswi2016movieqa} and its homepage\footnote{\url{http://movieqa.cs.toronto.edu/}.}.
% In the Video and Subtitle category, there are about 2 hours of video clips and subtitle texts aligned in time, and the remaining 4 story sources are composed of text only.

Among six tasks, we discuss our results with more focus on the video+subtitle task, because it is the only VQA task that requires both video and text understanding, whereas the other tasks are text-only. 
We weight less on the plot synopses only task, since plot synopses are given with a question, and all the QA pairs are generated from plot synopses, this task can be tackled using simple word/sentence matching algorithms (with little movie understanding), achieving a very high accuracy of 77.63\%. 

We solve the video+subtitle task using the proposed RWMN model in Figure \ref{fig:RWMN}.
For the four text-only QA tasks, no visual sources $\{v_1, ... ,v_n\}$ are given, 
thus we use $\{s_1, ... ,s_n\}$ only to construct the movie embedding $\mathbf{E}$ of Eq.(\ref{eq:cbp}) without the CBP.
Except this, we use the same RWMN model to solve four text-only QA tasks. 

%-----------------------------------------------------------------------------
% Table 2: Video-based MovieQA performance
\begin{table}
\centering
\small
\newcommand{\ranked}[1]{\xspace\scriptsize\sf{(#1)}}
\begin{tabular}{l|cc}
\hline
\multirow{2}{*}{Methods} & \multicolumn{2}{c}{Video+Subtitle} \\
& val & test \\
\hline
OVQAP & -- & 23.61 \\
Simple MLP & -- & 24.09 \\
LSTM + CNN & -- & 23.45 \\
LSTM + Discriminative CNN & -- & 24.32\\

VCFSM & -- & 24.09  \\
DEMN~\cite{kim2017deepstory} & -- & 29.97 \\
MEMN2N~\cite{tapaswi2016movieqa} & 34.20 & -- \\
\hline
RWMN-noRW & 34.20 & --  \\
RWMN-noR & 36.50 & -- \\
RWMN-noQ & 38.17 & -- \\
RWMN-noVid & 37.20 & -- \\
RWMN & \textbf{38.67} & \textbf{36.25} \\
RWMN-bag & 38.37 & 35.69 \\
RWMN-ensemble & 38.30 & -- \\
\hline
\end{tabular}
\vspace{6pt}
\caption{Performance comparison for the video+subtitle task on MovieQA public validation/test dataset.
(--) means that the method does not participate on the task. 
Baselines include DEMM (Deep embedded memory network), OVQAP (Only video question answer pairs) and VCFSM (Video clip features with simple MLP).}
\vspace{-5pt}
\label{tab:results_video}
\end{table}

%-------------------------------------------------------------------------------------------
%-----------------------------------------------------------------------------
% Table 3: Text-based MovieQA performance
\begin{table*}[t]
\begin{centering}
\small
\newcommand{\ranked}[1]{\xspace\scriptsize\sf{(#1)}}
\begin{tabular}{l|ccccccccc}
\hline
Method  & \multicolumn{2}{c}{Subtitle} & \multicolumn{2}{c}{Script} & \multicolumn{2}{c}{DVS} & \multicolumn{2}{c}{Plot Synopses} & Open-end \\
& val & test & val & test & val & test & val & test & test \\
\hline
MEMN2N~\cite{tapaswi2016movieqa} & 38.0 & 36.9 & 42.3 & 37.0 & 33.0 & \textbf{35.0} & 40.6 & 38.4 &--  \\
SSCB-W2V~\cite{tapaswi2016movieqa} & 24.8 & 23.7 & 25.0 & 24.4 & 24.8 & 24.9 & 45.1 & 45.6 & -- \\ 
SSCB-TF-IDF~\cite{tapaswi2016movieqa}  & 27.6 & 26.5 & 26.1 & 23.9 & 24.5 & 23.3 & 48.5 & 47.4 & --  \\ 
SSCB Fusion~\cite{tapaswi2016movieqa}  & 27.7 & -- & 28.7 & -- & 24.8 & -- & 56.7 & 56.7 & --\\
CNN Word Matching~\cite{wang2017compare}  & -- & -- & -- & -- & -- & -- & \textbf{72.1} & 72.9 & --\\
Convnet Fusion (TF-IDF + Word2Vec) & -- & -- & -- & -- & -- & -- & -- & \textbf{77.6} & -- \\
Longest Answer  & -- & -- & -- & -- & -- & -- & -- & -- & 25.6\\
	
\hline
{RWMN} & \textbf{40.4} & \textbf{38.5} & \textbf{44.0} & \textbf{39.4} & \textbf{40.0} & 34.2 & 37.0 & 34.8 & \textbf{36.6} \\
\hline
\end{tabular}
\vspace{6pt}
\caption{
Performance comparison for all the tasks on MovieQA public validation/test dataset.
(--) indicates that the method does not participate on the task. 
The description of baselines with no reference can be found in the MovieQA leaderboard. 
}
\label{tab:results_text}
\end{centering}
\end{table*}
%-----------------------------------------------------------------------------
%%%%%% 4.2 Baselines
\subsection{Baselines}
\label{sec:movieqa_dataset}

We compare the performance of our approach with those of all the methods proposed in the original MovieQA paper~\cite{tapaswi2016movieqa} or in the official MovieQA leaderboard\footnote{\url{http://movieqa.cs.toronto.edu/leaderboard/} as of the ICCV2017 submission deadline (March 27, 2017 23:59 GMT).}.
We describe the baseline names in the caption of each result table. 
% They include the three baseline models of the MovieQA paper~\cite{tapaswi2016movieqa}. 

\begin{comment}
\textbf{Searching Student} denoted by {\tt SS-*};
A model that matches the subset $S'_{movie}$ of story $S_movie$ to the most similar query $q$ and answer $a$.
They find a window with the highest cosine similarity for $S'_{movie}$ , $q$, and $a$ included in the window while moving a window of a certain size and define it as a score for the corresponding answer.

\textbf{Searching Student with a Convolution Brain} denoted by {\tt SSCB-*};
It is similar to the Searching Student, but the score is not calculated as window and cosine similarity but is trained through the convolution layer.
They construct two matrices with a dot product between $s$ and $a$ and between $s$ and $q$, then concatenate two matrices and learn the score through two convolution operations, mean and max pooling.

\textbf{End2End Memory network} denoted by {\tt MEMN2N}
In order to apply \cite{sukhbaatar2015end} to MovieQA dataset, they share both weight for constructing two memories and weight for obtaining query and answer embedding and use pretrained word embedding.
However, unlike our RWMN, they did not have a Write/Read Network, and they learned video and text embedding with image-sentence ranking model\cite{kiros2014unifying, zhu2015aligning}, not Compact BIlinear Pooling\cite{fukui2016multimodal}.
\end{comment}

% We also test several variants of our RWMN model. 
In order to measure the effects of key components of the RWMN, 
we experiment with five variants: 
(i) (RWMN-noRW) model without read/write networks, (ii) (RWMN-noR) model with only the write network, (iii) (RWMN-noQ) model without query-dependant memory embedding, (iv) (RWMN-noVid) model trained without using videos to quantify the importance of visual input, and (v) (RWMN) model with both write/read networks.
%is a subset of the subtitle-only task dataset as subtitles for the video+subtitle task exist only partially.

We also test two ensemble versions of our model.
Since the MovieQA dataset size is relatively small compared to task difficulty (\eg 4,318 training QA examples in video+subtitle category),
models often suffer from severe overfitting, which the ensemble methods can mitigate.
The first (RWMN-bag)  is a bagged version of our approach, in which we independently learn RWMN models on 30 bootstrapped datasets, and obtain the averaged prediction. 
The second (RWMN-ensemble) is a simple ensemble, in which we independently train 20 models with different random initializations, and compute the average prediction. 

\subsection{Quantitative Results}
\label{sec:quant_results}

% The results of this experiment are shown in the Table \ref{tab:results_movie}.
We below report the results of each method on the validation and test sets, both of which are not used for training at all. 
While the original MovieQA paper~\cite{tapaswi2016movieqa} reports the results on the validation set only,
the official leaderboard shows the performance on the test set only, 
for which groundtruth answers are not observable and the evaluation is performed through the evaluation server.
The test submission to the server is limited to once every 72 hours. 

%We perform 12 independent experiments with the same hyperparameter for each task and report the highest performance among them. The variance of performance is less than 1-2\% for all tasks.

As of the ICCV2017 submission deadline, our RWMN achieves the best performance for \textit{four} out of five tasks in the validation set, and \textit{four} out of six tasks in the test set.

\smallskip
\textbf{Results of VQA task}. 
Table \ref{tab:results_video} compares the performance of our RWMN model with those of baselines for the video+subtitle task.
We observe that RWMN achieves the best performance on both validation and test sets.
For example, in the test set, RWMN attains 36.25\%, which is significantly better than the runner-up DEMN of 29.97\%.

As expected, the RWMN with both read/write networks is the best among our variants on both validation and test sets.
It implicates that read/write networks play a key role in improving movie understanding.
For example, the RWMN-noR with only write network attains higher performance than the RWMN-noRW, which has similar or lower performance than other existing models. 
The RWMN-noQ without question-dependent memory embedding also underperforms the normal RWMN, which shows that the memory update according to the question is indeed helpful to select a more relevant answer to the question.
Finally, the RWMN-noVid is not as good as the RWMN, meaning that our RWMN successfully exploits both full videos and subtitles for training.
% to answer a question than training using only subtitles, and such multimodal information of videos and text is  understood by. 
Interestingly, the ensemble methods of our model, RWMN-bag and RWMN-ensemble, slightly underperform the  single model RWMN. 

\smallskip
\textbf{Results of text-only tasks}. 
Table \ref{tab:results_text} shows the results on the validation and test sets for text-only categories (\ie subtitle only, DVS only, script only, plot synopses only).
\begin{comment}
We defer the results on the test set to the supplementary material.
\end{comment}
For the open-end task, we simply use the plot synopses version of our method, which outperforms the only trivial baseline for the test set (\ie selecting the longest answer choice).

Our RWMN achieves the best performance in all tasks except for DVS-test set and plot synopses task.
\begin{comment}
In the script only task, our RWMN slightly underperforms the best baseline MEMN2N~\cite{tapaswi2016movieqa}, but the margin is trivial (\eg RWMN: 42.1 and MEMN2N : 42.3).
\end{comment}
We also observe that the ensemble methods hardly improve the performance of our method noticeably.
As discussed before, the memory network approaches including our RWMN and MEMN2N are not outstanding in the plot synopses only category.
It is mainly due to that the queries and answer choices are made directly from the plot sentences, 
and thus, this task can be tackled better by word/sentence matching methods with little story comprehension.
In addition, %as shown in Table \ref{tab:dataset}, 
each plot synopsis consists of about 35 sentences on average as a summary of a movie, which is much shorter than other data types,
for examples, about 1,558 sentences of subtitles per movie. 
Therefore, the memory abstraction by our method becomes less critical to solve the problems in this category. 

One important difference between the four text-only tasks is that each story source has a different $n$ (\ie the number of sentences), 
and thus the density of information contained in each sentence is also different. 
For example, the average $n$ of the scripts is about 2,877 per movie, while the average $n$ of DVS is about 636;
thus, each sentence in the script contains low-level details, while each sentence in the DVS contain high-level and abstract content. 
Given that the performance improvement by our RWMN is more significant in the DVS only task 
(\eg RWMN: 40.0 and MEMN2N: 33.0), 
it can be seen that our proposal to read/write networks may be more beneficial to understand and answer high-level and abstract content.

%As shown in Table \ref{tab:results_video} and \ref{tab:results_text}, 

%-------------------------------------------------------------------------------------------

%-------------------------------------------------------------------------------------------
%%%%%% 4.4 Ablation results
\subsection{Ablation Results}
\label{sec:ablation_results}

We experiment the performance variation according to the structure of CNNs in the write/read networks.
Among hyperparameters of the RWMN, the following three combinations have significant effects on the performance of the model;
i) conv-filter/stride sizes of the write network ($f^w_{v}, s^w_{v}, f^w_{c}$), ii) conv-filter/stride sizes of the read network ($f^r_{v}, s^r_{v}, f^r_{c}$), and iii) number of read/write CNN layers $\nu_r, \nu_w$.
Regarding the convolutions, the larger the convolution filter sizes, the more memories are read/written as a chunk.
Also, as the stride size decreases or the number of output channels increases, the total number of memory blocks increases.

Table \ref{tab:ablation} summarizes the performance variation on the video+subtitle task according to different combinations of these three hyperparameters. 
We make several observations from the results.
First, as the number of CNN layers in read/write network increases, the capacity of memory interaction may increase as well; yet the performance becomes worsen. 
Presumably, the main reason may be overfitting due to a  relative small dataset size of MovieQA as discussed. 
It is hinted by our results that the two-layer CNN is the best for training performance, while the one-layer CNN is the best for validation. 
% Even so, in almost all cases the Multi-Layer Write / Read Network outperformed the baseline.
Second, we observe that there is no absolute magic number of how many memory slots should be read/written as a single chunk and how many strides the memory controller moves. 
If the stride height is too small or too large compared to the height of a convolution filter, the performance decreases. 
It means that the performance can be degraded when too much information is read/written as a single abstracted slot, when too much information is overlapped in adjacent reads/writes (due to a small stride), or when the information overlap is too coarse (due to a high stride). 
We present more ablation results to the supplementary file.  
% The relationship between the convolution filter height and the stride height was further investigated in the 1-layer CNN setting and attached to the supplementary material.

\begin{figure}[t!]
\centering
\includegraphics[trim=0.2cm 0.2cm 0cm 0.1cm,clip,width=0.47\textwidth]{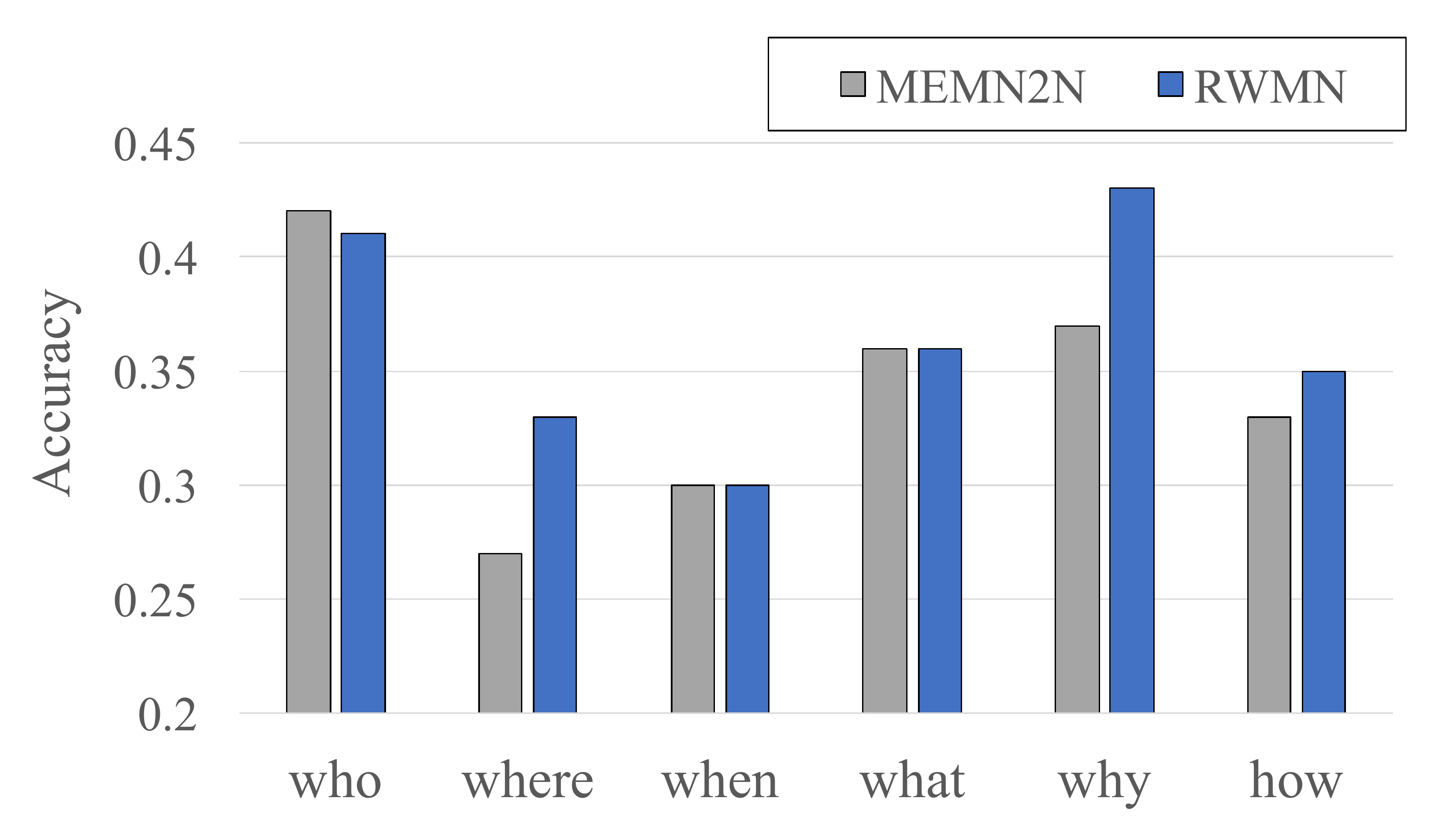}
\caption{Accuracy comparison between RWMN and the MEMN2N~\cite{tapaswi2016movieqa} baseline on the video+subtitle task according to question types. 
The RWMN leads higher improvement for \textit{Why} questions that often require abstract and high-level understanding.}
\label{fig:perf_qtype}
\end{figure}

Figure \ref{fig:perf_qtype} compares between the MEMN2N \cite{tapaswi2016movieqa} and our RWMN model according to question types in the video+subtitle task. 
We examine the results of six question types, according to what starting word is used in the question: \textit{Who, Where, When, What, Why}, and \textit{How}.
Usually, \textit{Why} questions require abstraction and high-level reasoning to answer correctly (\eg \textit{Why did Harry end his relationship with Helen?, Why does Michael depart for Sicily?}).
On the other hand, \textit{Who} and \textit{When} questions primarily deal with factual elements (\eg \textit{Who is Harry's girlfriend?, When does Grissom plan to set up Napier to be murdered?}).
Compared to the MEMN2N~\cite{tapaswi2016movieqa}, our RWMN shows higher performance enhancement in the questions starting with \textit{Why}, 
which may implicate the superiority of the RWMN to deals with high-level reasoning questions. 

%-------------------------------------------------------------------------------------------

%---------------------------------------------------------------------
%%%%%%% Table 4. 
\begin{table}[t]
\centering
\small
\setlength{\tabcolsep}{0.5em}
%\resizebox{.60\textwidth}{!}{
\begin{tabular}{cc|c|c|c}
\hline
\multicolumn{2}{c|}{\# Layers} & Write network  & Read network & Acc. \\
$\nu_w$ & $\nu_r$  & $(f^w_{vi}, s^w_{vi}, f^w_{ci})$  &  $(f^r_{vi}, s^r_{vi}, f^w_{ri})$ & \\
\hline
0 & 0 & -- & -- & 34.2\\
\hline
1 & 0 & (40,7,1) & -- & 33.9\\
1 & 0 & (40,30,3) & -- & 36.5\\ 
1 & 1 & (40,30,3) & (3,1,1) & \textbf{38.6}\\
1 & 1 & (40,60,3) & (3,1,1) & 33.6\\
\hline
2 & 1 & (40,10,3), (10,5,3) & (3,1,1) & 37.2\\
2 & 1 & (5,3,1), (5,3,1) & (3,1,1) & 37.3\\
2 & 2 & (4,2,1), (4,2,1) & (3,1,1), (3,1,1) & 36.9\\
2 & 2 & (4,2,1), (4,2,1) & (4,2,1), (4,2,1) & 37.3\\
\hline
3 & 1 & (10,3,3), (40,3,3), (100,3,3) & (3,1,1) & 35.1\\
3 & 1 & (40,3,3), (10,3,3), (10,3,3) & (3,1,1) & 37.9\\
3 & 1 & (40,3,3), (40,3,3), (40,3,3) & (3,1,1) & 35.7\\
3 & 1 & (100,3,3), (40,3,3), (10,3,3) & (3,1,1) & 35.8\\ 
%3 & 2 \\
%3 & 2 \\
\hline
\end{tabular}
%}
\vspace{6pt}
\caption{Performance of the RWMN on the video+subtitle task, according to the structure parameters of write/read networks.
$\nu_{w/r}$: the number of layers for write/read networks, $(f^{w/r}_{vi}, s^{w/r}_{vi}, f^{w/r}_{ci})$: the height and the stride of convolution filters, and the number of output channels.}
\label{tab:ablation}
\end{table}

%----------------------------------------------------------
%%%%%%%%% Qualitative results
\subsection{Qualitative Results}
\label{sec:qual_results}

%------------------------------------------------------------------------------
% Figure 3: question type performance
%------------------------------------------------------------------------------

\begin{figure*}[t!]
\centering
\includegraphics[width=0.99\textwidth]{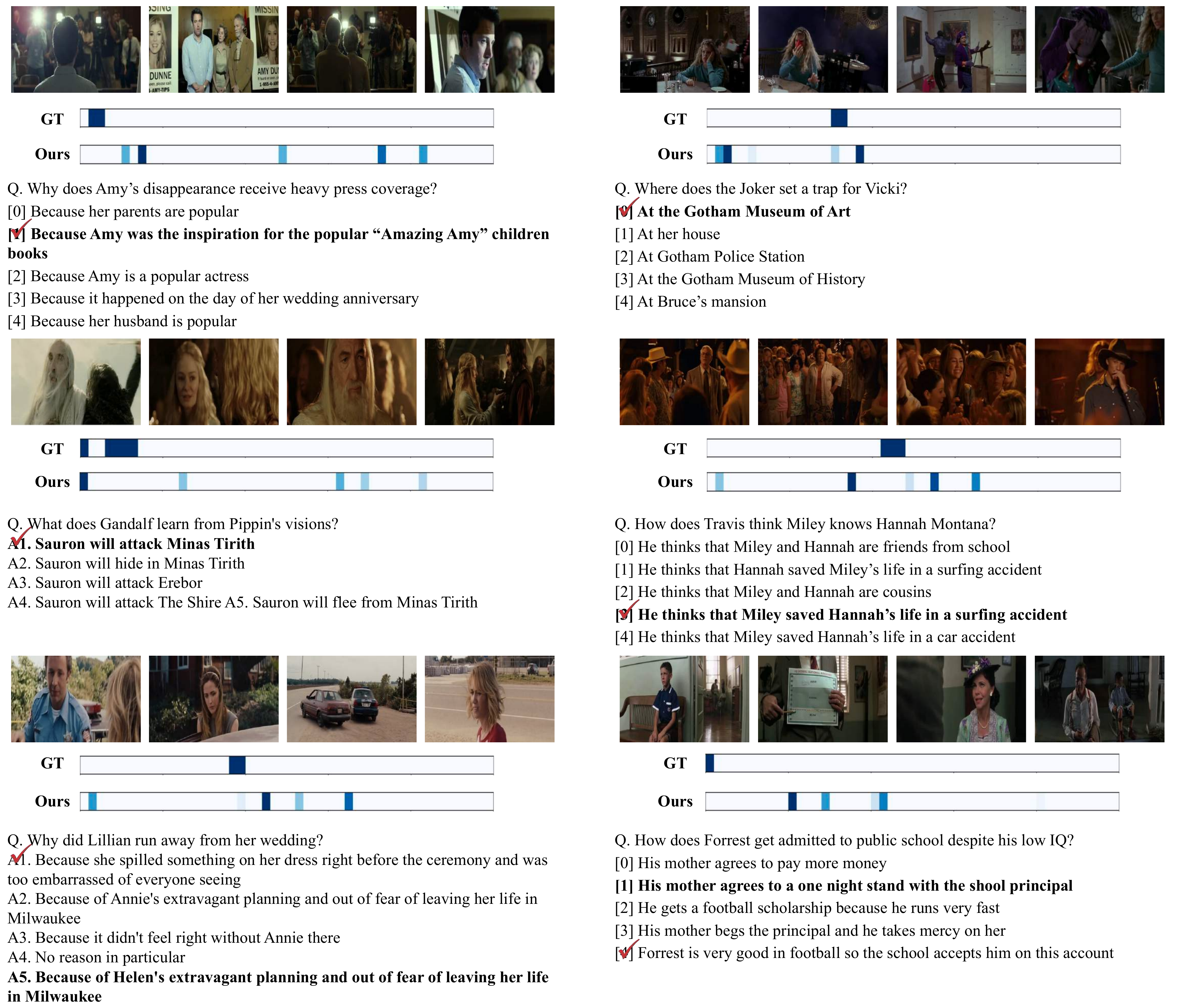}
\caption{Qualitative examples of MovieQA video+subtitle problems solved by our methods  (success cases in the top two rows, and failure cases in the last row). Bold sentences are groundtruth answers and red check symbols indicate our model's selection.
In each example, we also show on which parts our RWMN model attend over entire movie.
The attention by the RWMN often matches well with the groundtruth (GT) where the question is actually generated.}
\label{fig:attention}
\end{figure*}

Figure \ref{fig:attention} illustrates selected qualitative examples of video+subtitle problems solved by our methods, including four success and two near-miss cases. 
In each example, we present a sampled query video, a question, and five answer choices in which groundtruth is in bold and our model's selection is red checked.  
We also show on which parts our RWMN attends over entire movies, along with the groundtruth (GT) attention maps indicating the temporal locations of the clips where the question is actually generated, provided by the dataset. 
As examples show, movie question answering is highly challenging, and sometimes is not easy even for human. 

Our predicted attention often agrees well with the GT; the RWMN can implicitly learn where to place its attention in a very long movie for answering, although such information is not available for training. 
However, sometimes the RWMN can find correct answers even with the attention mismatch with the GT.
It is due to that  the MovieQA dataset also includes many questions that are hardly solvable with only attending on the GT parts.
That is, some questions require understanding the relationship between characters or progress of event development, for which attending beyond GT parts is necessary. % Therefore, QA performance is not always correlated with how we accurately localize the clip parts given by the dataset.

%------------------------------------------------------------------------
%%%%%%%%% 5. Conclusion
\section{Conclusion}

We proposed a new memory network model named Read-Write Memory Network (RWMN),
whose key idea is to propose the CNN-based read/write network that enable the model to have highly-capable and flexible read/write operations. 
We empirically validated that the proposed read/write networks indeed improve the performance of visual question answering tasks for large-scale, multimodal movie story understanding.
Specifically, our approach achieved the best accuracies in multiple tasks of MovieQA benchmark, with a significant improvement on visual QA task. 
We believe that there are several future research directions that go beyond this work. 
First, we can apply our approach to other QA tasks that require complicated story understanding. 
Second, we can explore better video and text representation methods beyond ResNet and Word2Vec. 

\textbf{Acknowledgements}.
This research is partially supported by SK Telecom and Basic Science Research Program through National Research Foundation of Korea (2015R1C1A1A02036562).
Gunhee Kim is the corresponding author.

{\small
\bibliographystyle{ieee}
\bibliography{iccv17_movieqa}
}

\end{document}